\documentclass[letterpaper, 10 pt, journal, twoside]{IEEEtran}  

\IEEEoverridecommandlockouts                              

\usepackage{fancyhdr,graphicx,amsmath,amssymb}
\usepackage[ruled,vlined]{algorithm2e}
\usepackage{tabu}

\usepackage{graphics} 
\usepackage{graphicx}
\usepackage{epsfig} 
\usepackage{amsmath} 
\usepackage{amssymb}  
\usepackage{cite}
\usepackage{multirow}
\usepackage{multicol}
\usepackage{caption}
\usepackage{acronym}
\usepackage{subcaption}

\let\labelindent\relax
\usepackage[shortlabels]{enumitem}
\usepackage{xr}

\usepackage[usenames,dvipsnames]{xcolor}

\usepackage{booktabs}
\usepackage{soul}

\DeclareMathAlphabet{\mbf}{OT1}{ptm}{b}{n}

\include{pythonlisting}

\graphicspath{{figs/}}

\author{Yuxuan Chen$^{1}$, Timothy D. Barfoot$^{1}$%
\thanks{Manuscript received: August, 26, 2022; Revised November, 2, 2022; Accepted November, 29, 2022.}
\thanks{This paper was recommended for publication by Editor Sven Behnke upon evaluation of the Associate Editor and Reviewers' comments. We thank Mona Gridseth for her tremendous help with this project.}
\thanks{$^{1}$Yuxuan Chen and Timothy D. Barfoot are with the University
of Toronto Institute for Aerospace Studies, University of Toronto,
Canada \texttt{\{yuxuansherry.chen, tim.barfoot\}@mail.utoronto.ca}}%
\thanks{Digital Object Identifier (DOI): see top of this page.}
}

\title{
Self-Supervised Feature Learning for Long-Term Metric Visual Localization
}

\begin{document}

\maketitle


\begin{abstract}
Visual localization is the task of estimating camera pose in a known scene, which is an essential problem in robotics and computer vision. However, long-term visual localization is still a challenge due to the environmental appearance changes caused by lighting and seasons. While techniques exist to address appearance changes using neural networks, these methods typically require ground-truth pose information to generate accurate image correspondences or act as a supervisory signal during training. In this paper, we present a novel self-supervised feature learning framework for metric visual localization. We use a sequence-based image matching algorithm across different sequences of images (i.e., experiences) to generate image correspondences without ground-truth labels. We can then sample image pairs to train a deep neural network that learns sparse features with associated descriptors and scores without ground-truth pose supervision. The learned features can be used together with a classical pose estimator for visual stereo localization. We validate the learned features by integrating with an existing Visual Teach \& Repeat pipeline to perform closed-loop localization experiments under different lighting conditions for a total of 22.4 km.
\end{abstract}



\section{INTRODUCTION}
\vspace{-0.1cm}
\IEEEPARstart{L}{ong-term} navigation across drastic appearance change is a challenge in visual localization. Traditional point-based localization approaches find correspondences between local features extracted from images by applying hand-crafted descriptors (e.g., SIFT, SURF, ORB~\cite{SIFT, SURF, ORB}), then recover the full 6-DoF camera pose. However, such hand-crafted features are not robust under extreme appearance changes. To address this, Multi-Experience Visual Teach \& Repeat\cite{vtr, MEL}, a visual-based route-following framework, proposed to store intermediate experiences to achieve long-term localization. 

Due to recent developments in the field of deep learning, many recent works use neural networks to directly predict relative poses~\cite{relocnet, CamNet, NN_Net} or absolute poses~\cite{PoseNet} from images for localization. Instead of directly learning poses from images, deep-learned interest-point detectors and descriptors \cite{mona_vision, D2Net, TILDE, superpoint, LIFT, LFNet, GN_Net, R2D2, DarkPoint, unsup_metric_reloc_trfm_csty} have gained popularity since they produce more accurate results than direct pose regression when combined with a classical pose estimator. In addition, Gridseth and Barfoot \cite{mona_vision} have proven the effectiveness of deep-learned features in the VT\&R framework under different illumination conditions. However, these feature learning networks typically require training with accurate image correspondences that are extracted from known scene geometry or ground-truth camera poses. Although some methods simplify the problem by collecting data using a stationary camera~\cite{TILDE}, applying known homographic adaptation\cite{superpoint, R2D2, DarkPoint}, or rendering synthetic training data\cite{superpoint, GN_Net}, the generalizability on unseen data is limited. 

In this paper, we propose a novel self-supervised feature learning framework for long-term visual localization that does not require any ground-truth labels while preserving generalizability. We identify that one of the main challenges is to find accurate image correspondences in unstructured outdoor environment without ground-truth data. Since the ultimate goal is to learn features in a self-supervised manner, we do not want to presuppose the existence of local features. Especially, we want to avoid using deep-learned features that are pretrained on other datasets using known image correspondences. Hence, we adopt a sequence-based visual place recognition algorithm that works directly with whole-image descriptors to generate image pairs. Our overall pipeline consists of two stages: place recognition and feature learning. During the first stage, we generate training samples by performing sequence-based visual place recognition on data collected under different seasons and lighting conditions. During the second stage, we train a feature learning network using the sampled training stereo image pairs to predict keypoints with associated descriptors and scores in an Expectation-Maximization loop without any ground-truth pose supervision.

\vspace{-0.1cm}

\section{RELATED WORK}
\vspace{-0.1cm}

\begin{figure}[t]
	\centering
	\includegraphics[width=0.95\linewidth]{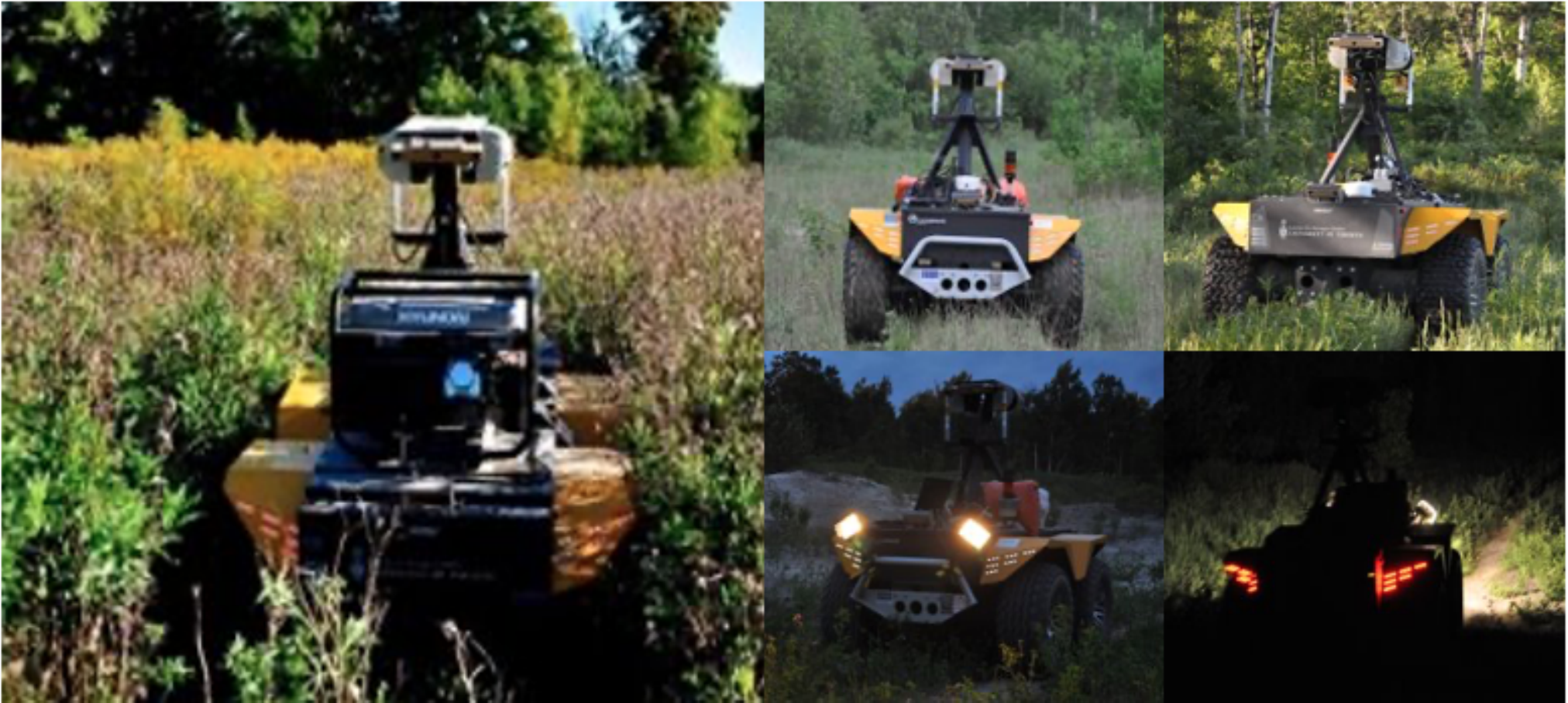}
	\caption{\textbf{Grizzly ground robot} autonomously repeats a taught path with learned features despite severe lighting changes.}
	\label{fig:grizzly}
	\vspace{-0.3cm}
\end{figure} 

\begin{figure*}
  \centering
    \includegraphics[width=0.9\textwidth]{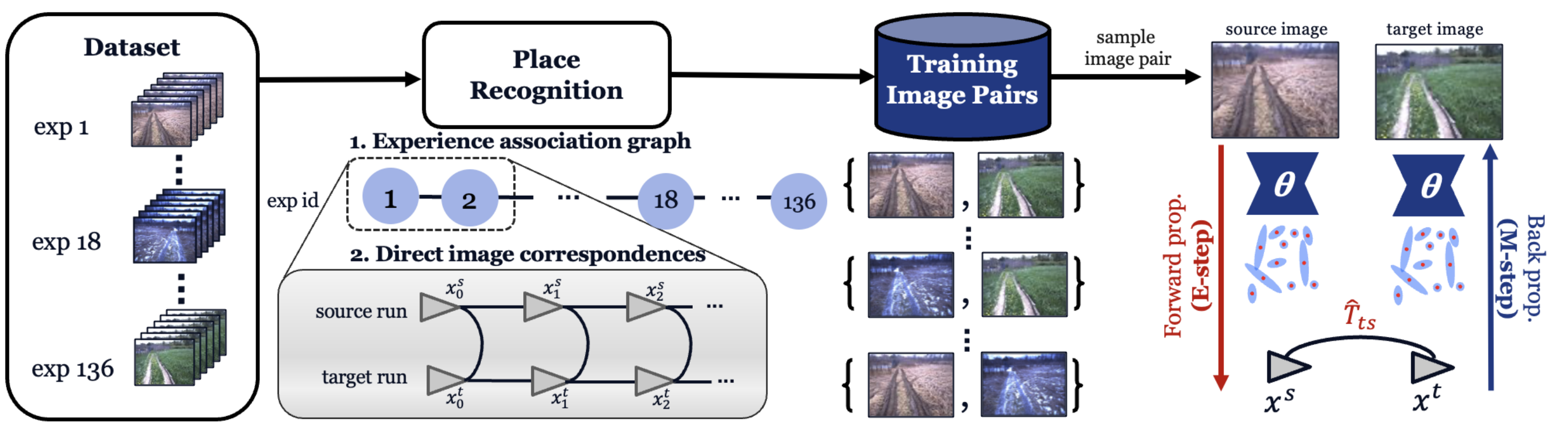}
    \caption{\textbf{Overall pipeline}. The dataset consists of multiple sequences (i.e., experiences) of image data organized by their collection times. During the place-recognition stage, we directly match neighbouring experiences using a place-recognition algorithm, and find indirect image correspondences by constructing an experience-association graph. We sample image pairs using the predicted image correspondences for the subsequent feature-learning stage. During the E-step of feature learning, the network predicts keypoints from the source and target images, which can be used to compute the relative transform. During the M-step, we use the estimated relative transform as a supervisory signal to optimize the network weights.}
    \label{fig:pipeline}
\end{figure*}

Traditionally, hand-crafted features have been commonly used for visual localization~\cite{SIFT, SURF, ORB}, but suffer from low repeatability on images under dramatic appearance changes\cite{sift_surf_seasons}. Experience-based visual navigation systems\cite{exp_based_long_term_loc, work_smart_not_hard,MEL} attempt to bridge significant appearance changes by storing multiple appearances of the same location (i.e., experiences) and choosing the most relevant experiences for feature matching during online operation.

To address the deficiencies of hand-crafted features, there has been a wide range of deep-learning-based approaches for pose estimation. Some methods tackle camera localization by training a CNN that directly learns relative poses\cite{NN_Net, relocnet, CamNet} or absolute poses\cite{PoseNet} from images. However, learning pose directly from image data can struggle with accuracy\cite{CNN_limit}.

Other works focus on only learning the visual features or descriptors\cite{under_the_radar, mona_vision, D2Net, TILDE, superpoint, LIFT, LFNet, GN_Net, R2D2, DarkPoint, unsup_metric_reloc_trfm_csty}, which can be integrated with a classical pose estimator for localization. For instance, Gridseth and Barfoot~\cite{mona_vision} and Sun et al.~\cite{DarkPoint} have proven the effectiveness of deep-learned features against illumination changes by integrating with the VT\&R pipeline for long-term visual localization. However, training networks for visual localization generally requires accurate image correspondences extracted from ground-truth camera poses or known scene geometry. Although Kasper et al.~\cite{unsup_metric_reloc_trfm_csty} proposed an unsupervised transform consistency loss to train features, it still requires GPS to ensure the query and reference images have sufficient visual overlap. Alternatively, Verdie et al.~\cite{TILDE} use a stationary camera to observe the same scene under different illumination conditions, whereas other methods obtain image correspondences from either using synthetically rendered data\cite{superpoint, GN_Net} or applying known transformations (e.g., colour shift, homographic transform) to real-world images\cite{superpoint, R2D2, DarkPoint}. However, these approaches limit the generalizability of the learned model on unseen data. 

In this paper, we propose to find image correspondences using a visual place-recognition algorithm without any labels. We will discuss some relevant place-recognition approaches in the rest of the section. Since our goal is to sample training data for feature learning in a self-supervised manner, we restrict our discussion to hand-crafted algorithms that do not involve any learning or human annotations. 

Lowry et al.~\cite{VPR_survey} provide a thorough overview of existing methods for place recognition. Traditional bag-of-words-based FAB-MAP2\cite{fabmap} shows good performance, but tends to fail in the presence of dramatic appearance changes\cite{app_chg_pred_long_term_nav} due to the limited repeatability of local hand-crafted keypoints\cite{SIFT, SURF} in changing environments\cite{sift_surf_seasons}. In order to improve the performance of place recognition under appearance changes, various approaches have been developed to exploit the sequential nature of the data \cite{work_smart_not_hard, howlowcanugo, seqSLAM}. SeqSLAM\cite{seqSLAM} is one of the most recognized sequence-based algorithms that achieves significant performance improvements over FAB-MAP2 under appearance changes. SeqSLAM finds image correspondences by constructing a pairwise similarity matrix between the local query image sequence and a database image sequence without the need for keypoint extraction. Hansen et al.~\cite{VPR_HMM} further improve the overall flexibility of the sequence alignment procedure by using a hidden Markov model (HMM) to formulate a graph-based sequence search method in the similarity matrix. In order to match image sequences under severe seasonal changes more efficiently, Naseer et al.~\cite{VPR_network_flows} formulate sequence matching as a minimum cost flow problem in an offline data-association graph. Vysotska et al.~\cite{lazy_data_assoc_img_seq_app_change} extended this idea to build a data-association graph online using a pretrained CNN for feature extraction, and later proposed a hashing-based relocalization strategy~\cite{VPR_multi_seq_map} to improve data association for flexible trajectories. Additionally, Neubert et al.~\cite{exploit_intra_db_sim} detect similarities within the database to avoid unnecessary image matching between query and database images. 

For our purposes, we adopt SeqSLAM aided by Visual Odometry (VO) to find direct image correspondences between temporally neighbouring experiences, and indirectly link further experiences by building an offline experience-association graph.

\vspace{-0.1cm}

\section{METHODOLOGY}
\label{methodology}
\vspace{-0.1cm}

We propose a two-stage pipeline as shown in Figure \ref{fig:pipeline}, which includes: 1) place recognition and 2) self-supervised feature learning. During the place-recognition stage, we build a locally connected experience-association graph by matching neighbouring experiences with gradual perceptual changes, where we can sample image pairs from direct or indirect image correspondences between experiences using a graph-search algorithm. Our fully differentiable feature-learning pipeline takes a pair of source and target stereo images generated from the previous stage, and estimates the relative pose, $\textbf{T}_{ts}\in SE(3)$, between their corresponding frames. During the feature-learning stage, we train a neural network in an Expectation-Maximization loop. In the E-Step, the network predicts sparse local keypoints with associated descriptors and scores for both input images. We find point correspondences between the predicted keypoints from the source and target images, which are used in a differentiable pose estimator to estimate the relative pose change. In the M-step, we construct a self-supervised keypoint loss using the estimated relative keypoint loss as a supervisory signal to optimize the network weights. 

    \begin{figure}[t]
        \vspace{+0.25cm}
    	\centering
    	\includegraphics[width=0.85\linewidth]{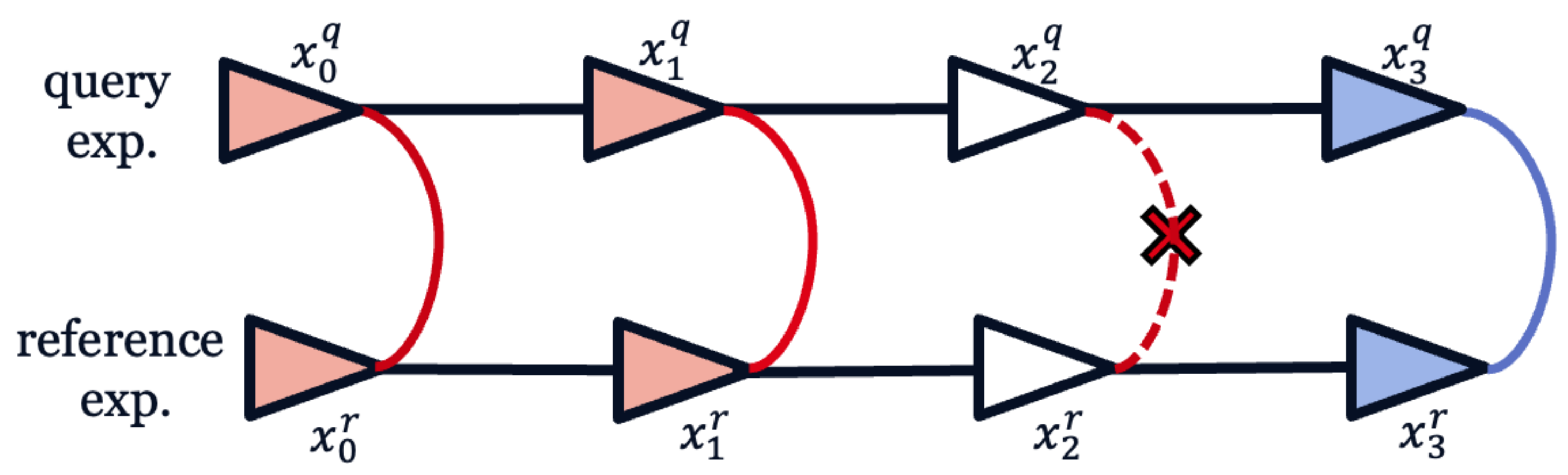}
    	\caption{\textbf{Direct image correspondences}. The black horizontal edges represent the relative transforms between frames (known from VO). The \textcolor{red}{red} vertical edges represent candidate matches suggested by SeqSLAM. The red solid edges are validated by VO, whereas the dotted edges are rejected by VO. After the candidate match is rejected, VO can also potentially find a new match in the reference sequence that is closest to the query image, which is represented by a \textcolor{blue}{blue} edge.}
    	\label{fig:frame_level_matching}
    	\vspace{-0.2cm}
    \end{figure} 

\subsection{Place Recognition}
\vspace{-0.1cm}

We define the set of training data $\mathcal{D} = (s_0, ..., s_M)$ as a temporally ordered set of $M$ experiences, where each experience is defined as a sequence of $N$ consecutive images $s_i = (x^i_0, ..., x^i_N)$. The ultimate goal is to generate direct or indirect image correspondences between any two experiences in the dataset $\mathcal{D}$. However, matching images of the same place under appearance change is a non-trivial task, as the appearance of the same place is likely to change substantially due to different illumination conditions and seasonal changes. A naive approach is to directly match images between experiences using proximity provided by Visual Odometry (VO). However, this results in poor image correspondences for longer routes due to drift errors in VO. Alternatively, we can exploit the sequential nature of the data and use SeqSLAM\cite{seqSLAM} to match different image sequences using patch-normalized image representation, but face two main problems:
\begin{enumerate}[a)]
    \item SeqSLAM is capable of matching experiences with gradual appearance changes but struggles with significant appearance gap.
    \item The raw SeqSLAM matching results may contain outliers and discontinuities.
\end{enumerate}

To mitigate these issues, we combine SeqSLAM with VO to improve the direct matching results, and construct an experience-association graph to extract indirect image correspondences between experiences. To sufficiently limit scope, we make the following assumptions:
\begin{enumerate}[a)]
    \item All experiences roughly follow the same trajectory with the same starting and ending positions. 
    \item All experiences are temporally ordered by their collection times. Nearby experiences are collected within a relatively short time frame, thus having similar appearances compared to further experiences. 
\end{enumerate}

    \begin{figure}[t]
        \vspace{+0.25cm}
    	\centering
    	\includegraphics[width=0.85\linewidth]{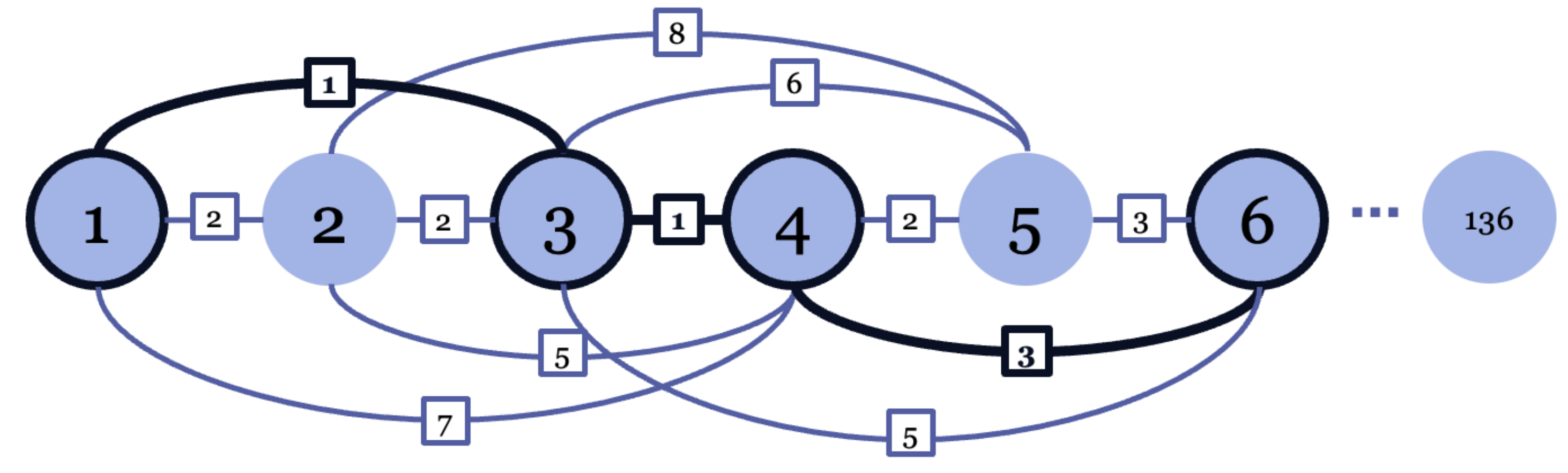}
    	\caption{\textbf{Experience-Association Graph}. Each vertex represents an experience, which is locally connected to 3 previous neighbours. and each edge represents direct image correspondences. The assigned cost term on each edge reflects the quality of the match. To compute indirect image correspondences between experience 1 and 6, the minimum cost path is indicated in black.}
    	\label{fig:multi_exp_mathing}
    	\vspace{-0.2cm}
    \end{figure} 
    

\subsubsection{Direct Image Correspondences}
According to the second assumption, we assume the temporally neighbouring experiences are collected within a shorter time frame, such that the appearance changes are small enough for us to find direct image correspondences using SeqSLAM and VO. For each query experience $s_i$, it is directly matched with $k$ previous experiences $\{s_{i-k}, ..., s_{i-1}\}$, which are denoted as the reference experiences. 

\begin{figure}[t]
    \vspace{+0.25cm}
    \centering
    \includegraphics[width=\linewidth]{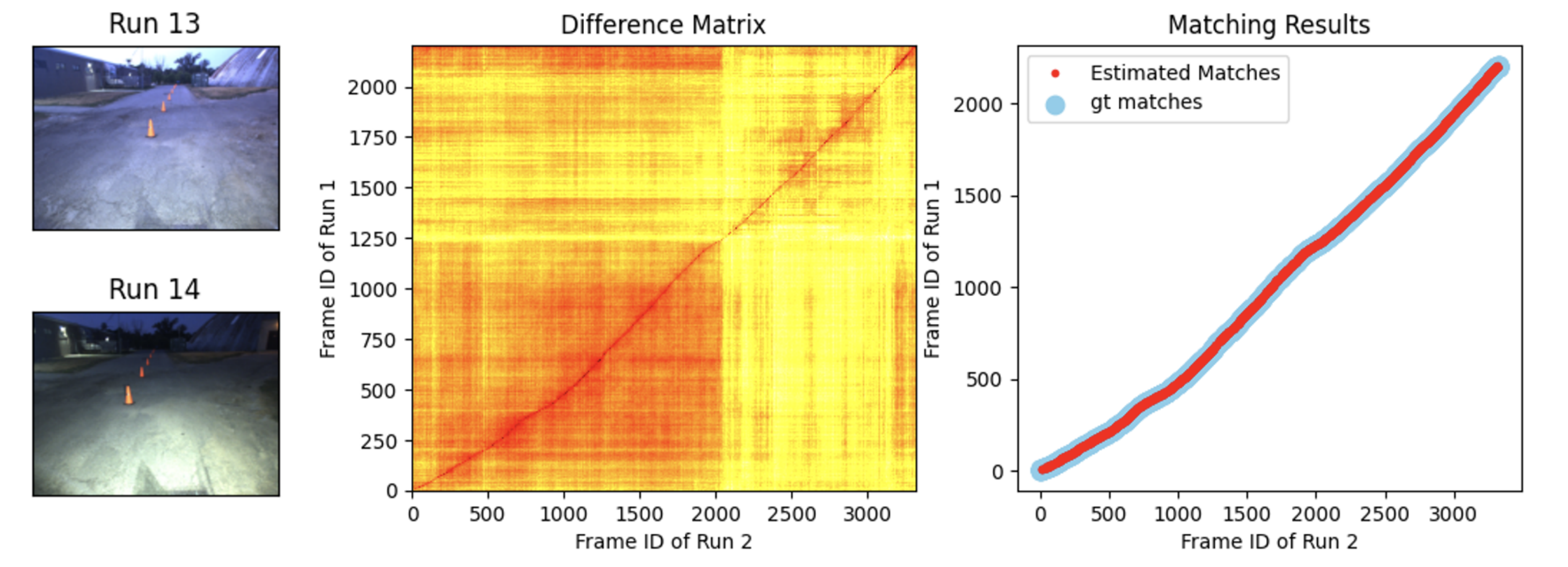}
    	\caption{\textbf{Frame-level topological matching results} between two neighbouring runs, run 13 and run 14, in the UTIAS-In-the-Dark dataset. The raw difference matrix generated by SeqSLAM algorithm and the processed matching results is shown.}
    	\label{fig:difference_matrix}
    	\vspace{-0.2cm}

\end{figure}

As shown in Figure~\ref{fig:difference_matrix}, SeqSLAM constructs a difference matrix by computing image-by-image dissimilarity scores between the two sequences using full-image descriptors, then finds raw image matches through the full matching matrix with the smallest sum of dissimilarity scores. We further validate the matching results using VO to reject outliers as illustrated in Figure~\ref{fig:frame_level_matching}.

For each image in the query experience, we use the raw SeqSLAM matching results to suggest a candidate match in the reference experience, then validate the match using VO. Assuming we have successfully matched and validated the image pair $\{q_0, r_0\}$ using SeqSLAM and VO, we consider the image pair as the last validated match $m$, then proceed to the next image in the query experience, $q_1$. After retrieving the candidate match $r_1$ from the reference experience using SeqSLAM, we use VO to estimate the relative transform between the last validated image and the current image in the query and reference experiences, namely $\hat{\textbf{T}}_{q_1, q_0}$ and $\hat{\textbf{T}}_{r_1, r_0}$. Since the last validated image pair $m=\{q_0, r_0\}$ are two images of the same place, we can approximate the relative transform $\hat{\textbf{T}}_{r_0, q_0}$ with an identity matrix $\textbf{I}$. As a result, the relative transform between the query and reference images can be estimated as:
\begin{multline}
    \hat{\textbf{T}}_{r_1, q_1} = \hat{\textbf{T}}_{r_1, r_0} \hat{\textbf{T}}_{r_0, q_0}\hat{\textbf{T}}_{q_1, q_0}^{-1} \\ 
    \approx \hat{\textbf{T}}_{r_1, r_0}\hat{\textbf{T}}_{q_1, q_0}^{-1} =  \begin{bmatrix}
        \textbf{C}_{r_1, q_1} & \textbf{r}_{r_1, q_1} \\ 
         \textbf{0}^T & 1 
\end{bmatrix}.
\end{multline}
The absolute distance between $q_1$ and $r_1$ can be computed as:
\begin{equation}
d_{r_1, q_1} = \left\| \textbf{r}_{r_1, q_1} \right\|^2_2.
\end{equation}
We compare the distance $d_{r_1, q_1}$ with a pre-specified threshold value $e$ to check if the candidate match is consistent with VO. We discuss two emerging cases as follows:
\begin{enumerate}[(a)]
    \item $d_{r_1, q_1} \leq e$. The candidate match suggested by SeqSLAM is validated by VO. Hence, we update the last validated match $m$ to be $\{q_1, r_1\}$ when processing the next image in the query sequence. 
    \item $d_{r_1, q_1} > e$. The candidate match suggested by SeqSLAM is rejected by VO. As a replacement, we retrieve a new image $r_j$ from the reference experience such that the distance between $r_j$ and $q_1$ is minimized. Note that we do not update the last validated match $m$ in this case.
\end{enumerate}

Since the last validated match $m$ is frequently updated, we usually only need to compute VO for a relatively short time frame starting from the last validated match. Hence, the matching results are less likely to be affected by VO drift.

\subsubsection{Indirect Image Correspondences}
According to the second assumption, relatively similar experiences are temporally closer together, whereas dissimilar experiences are further apart. Hence, we can find image correspondences between further experiences indirectly using bridging experiences. After finding direct image correspondences between neighbouring experiences, we can build a locally connected graph $G = \{V, E\}$, where $V$ is a set of vertices representing all experiences in the dataset, and $E$ is a set of edges representing direct image correspondences between the vertices. Each vertex is connected to $k$ previous vertices, and Figure \ref{fig:multi_exp_mathing} shows an example when $k=3$. 

We assign a cost term to each edge in $E$ to indicate the quality of the match. The cost term is defined as the number of frames rejected by VO out of the total number frames in the source sequence, where a higher cost represents a poorer match and a lower cost represents a better match. To generate image correspondences between two experiences that are not directly connected, we can find the minimum-cost path in the graph using a graph-search algorithm to indirectly extract image correspondences.

\subsection{Feature Learning}
\vspace{-0.1cm}

\subsubsection{Network Architecture}
Our network is adapted from the architecture presented by \cite{under_the_radar, mona_vision}, which is a U-Net style convolutional encoder-multi-decoder architecture to output keypoints, descriptors, and scores based on visual inputs. The encoder is a VGG16 network~\cite{vgg16} pretrained on the ImageNet dataset~\cite{IMGNet}, truncated after the $conv\_5\_3$ layer. 

The keypoint-locations decoder predicts the sub-pixel locations of each sparse 2D keypoint, $\textbf{q}=[u_l, v_l]^T$, in the left stereo image. To achieve this, we equally divide the image into $16 \times 16$ square cells, with each generating a single candidate keypoint. We then apply a spatial softmax on each cell and take a weighted average of the pixel coordinates to return the sub-pixel keypoint locations. 

In addition, the network computes the scores by applying a sigmoid function to the output of the second decoder branch. The scores $s$ are mapped to $[0,1]$, which predict how useful a keypoint is for pose estimation. Finally, we generate dense descriptors for each pixel by resizing and concatenating the output of each encoder layer, which results in a descriptor vector, $\mathbf{d} \in \mathbb{R}^{960}$.

\subsubsection{E-Step: Pose Estimation}
After generating $N$ keypoint predictions for the source image, we need to perform data association between the keypoints in the source and target images by performing a dense search for optimum keypoint locations in the target image. For each keypoint in the source image, we compute a matched point in the target image by taking the weighted sum of all image coordinates in the target image as follows:
\begin{equation}
    \label{eqn:kp_matching}
    \hat{\textbf{q}}^i_t = \sum^M_{j=1} \sigma(\tau f_{\rm zncc}(\textbf{d}_s^i, \textbf{d}_t^j))\textbf{q}_t^j,
\end{equation}
where M is the total number of pixels in the target image, $f_{\rm zncc}(\cdot)$ computes the zero-normalized cross correlation (ZNCC) between the descriptors, and $\sigma(\cdot)$ takes the temperature-weighted softmax with $\tau$ as the temperature. Finally, we find the descriptor, $\hat{\textbf{d}}_t^i$, and score, $\hat{s}^i_t$, for each computed target keypoint using bilinear interpolation. 

For the matched 2D keypoints, we compute their corresponding 3D coordinates using an inverse stereo camera model. The camera model, $\textbf{g}(\cdot)$, maps a 3D point, $\textbf{p}=[x,y,z]^T$, in the camera frame to a left stereo image coordinate, $\textbf{q}$, as follows:

\begin{equation}
    \label{eqn:stereo_model}
    \mbf{y} = \begin{bmatrix} u_l\\ v_l\\ d\end{bmatrix} = \begin{bmatrix}\textbf{q}\\ d
\end{bmatrix} = \textbf{g}(\textbf{p}) = \begin{bmatrix}
f_u & 0 & c_u &0 \\ 
0 & f_v & c_v & 0\\ 
0 & 0 & 0 & f_ub
\end{bmatrix}\frac{1}{z}\begin{bmatrix}
x\\ y\\ z\\ 1
\end{bmatrix},
\end{equation}
where $f_u$ and $f_v$ are the horizontal and vertical focal lengths in pixels, $c_u$ and $c_v$ are the camera's horizontal and vertical optical center coordinates in pixels, $d=u_l - u_r$ is the disparity obtained from stereo matching, and $b$ is the baseline in metres. We use the inverse stereo camera model to get each keypoint's 3D coordinates:
\begin{equation}
    \label{eqn:inverse_model}
    \textbf{p} = \begin{bmatrix}
x\\ 
y\\ 
z
\end{bmatrix} = \textbf{g}^{-1}(\mbf{y})=\frac{b}{d}\begin{bmatrix}
u_l-c_u\\ 
\frac{f_u}{f_v}(v_l-c_v)\\ 
f_u
\end{bmatrix}.
\end{equation}
Given the corresponding 2D keypoints $\{\textbf{q}_s^i, \hat{\textbf{q}}_t^i\}$ from the source and target images, we use (\ref{eqn:inverse_model}) to compute their 3D coordinates, $\{\textbf{p}_s^i, \hat{\textbf{p}}_t^i\}$. In addition, their descriptors and scores are: $\{\textbf{d}_s^i, \hat{\textbf{d}}_t^i\}$, and $\{s_s^i, \hat{s}_t^i\}$.

We then perform RANSAC to reject outliers prior to pose estimation to find features that are geometrically consistent with each other. This step is essential to provide a reasonable guess of the relative pose in the beginning of training to help the algorithm converge.

Given the inlier 3D keypoints, we can estimate the relative pose from the source to the target by minimizing the following cost using Singular Value Decomposition (SVD):

\begin{equation}
J = \sum^N_{i=1} w^i  \left \| (\textbf{C}_{ts}\textbf{p}^i_s + \textbf{r}^{st}_t) - \hat{\textbf{p}}^i_t  \right \|^2_2,
\end{equation}
where the weight for a matched point pair is a combination of the learned point scores and how well the descriptors match:
\begin{equation}
    w^i = \frac{1}{2} (f_{\rm zncc}(\textbf{d}_s^i, \hat{\textbf{d}}^i_t) + 1)s^i_s \hat{s}^i_t .
\end{equation}
Finally, we obtain the relative transform between the source and target images as:
\begin{equation}
\hat{\textbf{T}}_{ts} = 
    \begin{bmatrix}
    \textbf{C}_{ts} & \textbf{r}^{st}_t \\ 
     \textbf{0}^T & 1 
    \end{bmatrix}.
\end{equation}

\subsubsection{M-Step: Feature Optimization}
Barnes and Posner~\cite{under_the_radar} and Gridseth and Barfoot~\cite{mona_vision} use the ground-truth transform between the source and target images to construct a supervised pose loss. Since we do not rely on any ground-truth data, we directly use the estimated transform as a supervisory signal to construct a keypoint loss without any ground-truth pose information. We transform the predicted source keypoints $\textbf{p}_s^i$ to the target frame using the estimated relative transform $\hat{\textbf{T}}_{ts}$, and define a keypoint loss as follows:

\begin{equation}
    \label{eqn:keypoint_loss}
    \mathcal{L} = \sum^N_{i=1}{||\hat{\textbf{T}}_{ts} \textbf{p}_s^i - \hat{\textbf{p}}_t^i ||^2_2} .
\end{equation}

\begin{figure}[t]
    \centering
    \includegraphics[width=0.6\linewidth]{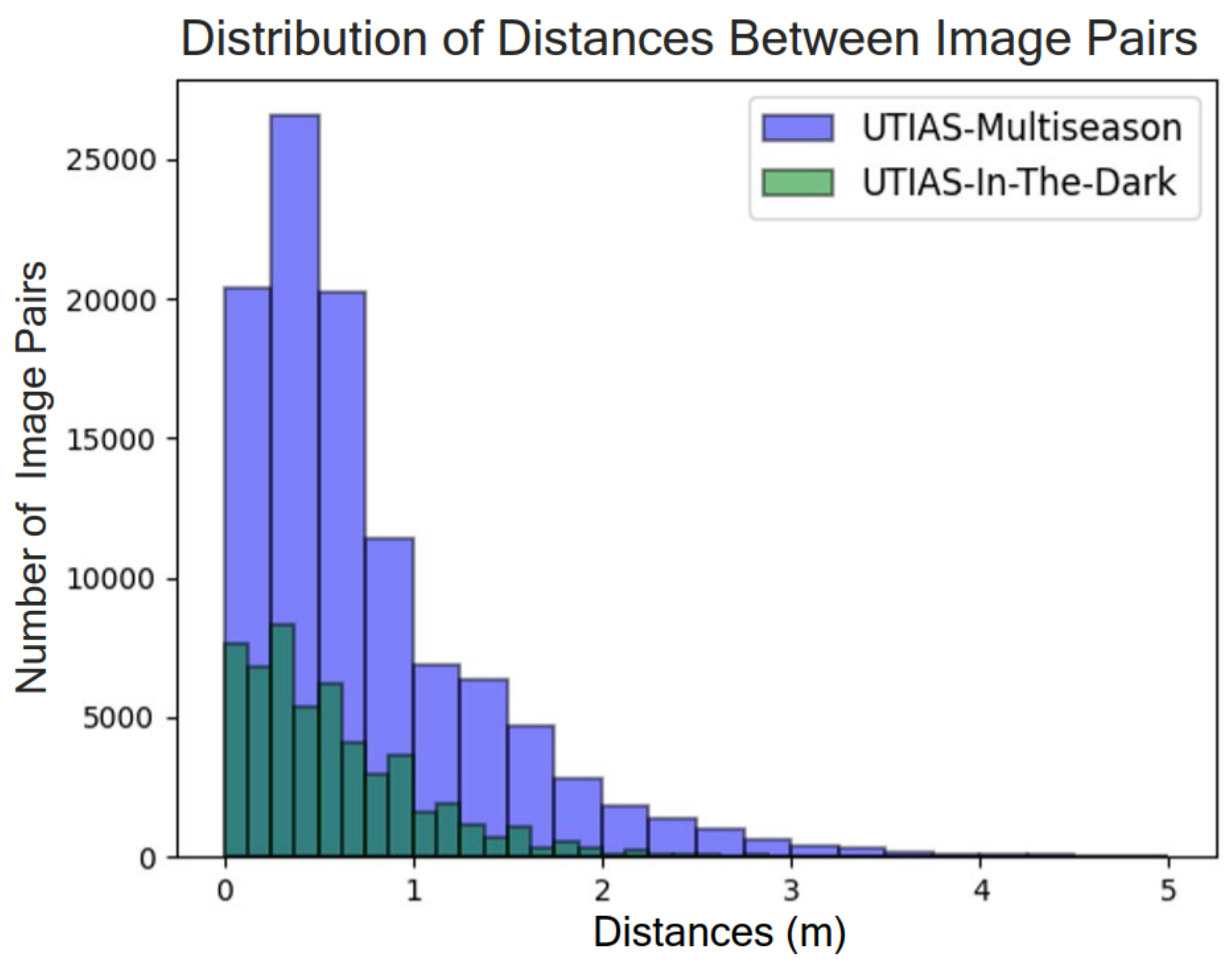}
    	\caption{\textbf{Distribution of distances between sampled image pairs} of Multiseason and In-the-Dark datasets.}
    	\label{fig:seqslam_training_testing_data}
    	\vspace{-0.2cm}
\end{figure}

\begin{figure}[h]
    \vspace{+0.25cm}
    \centering

      \includegraphics[width=0.95\linewidth]{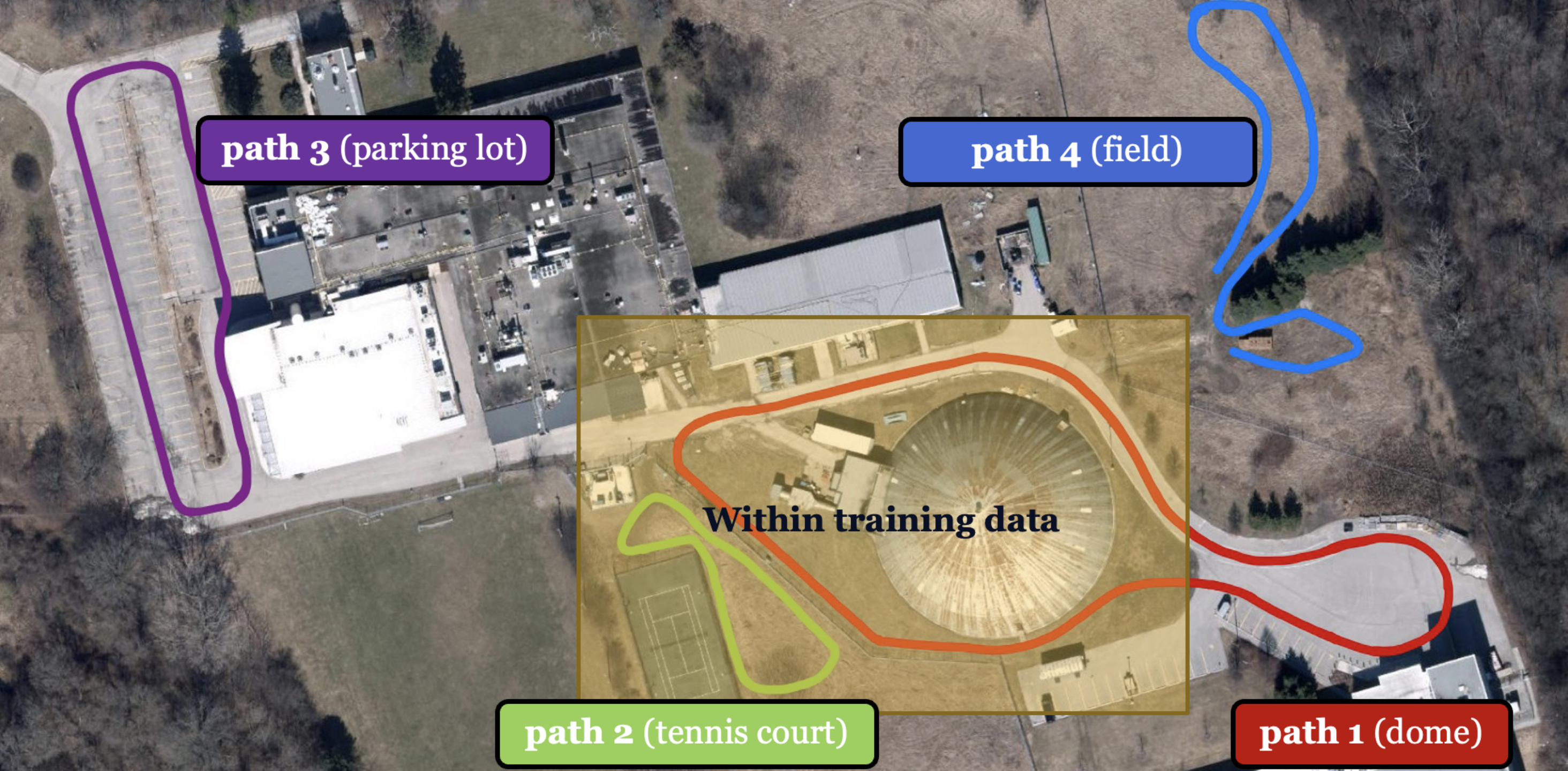}

    \caption{\textbf{Training and testing paths}. The training paths from the datasets are highlighted by the yellow region. \textcolor{green}{Path 1 (tennis court)} is collected in the same region as the training data, whereas the bottom right portion of \textcolor{red}{path 2 (dome)}, \textcolor{violet}{path 3 (parking lot)} and \textcolor{blue}{path 4 (field)} are not in training data.}
    \label{fig:training_testing_path}
\end{figure}

\begin{figure}[h!]
        \centering
        \vspace{-0.25cm}
      \includegraphics[width=0.95\linewidth]{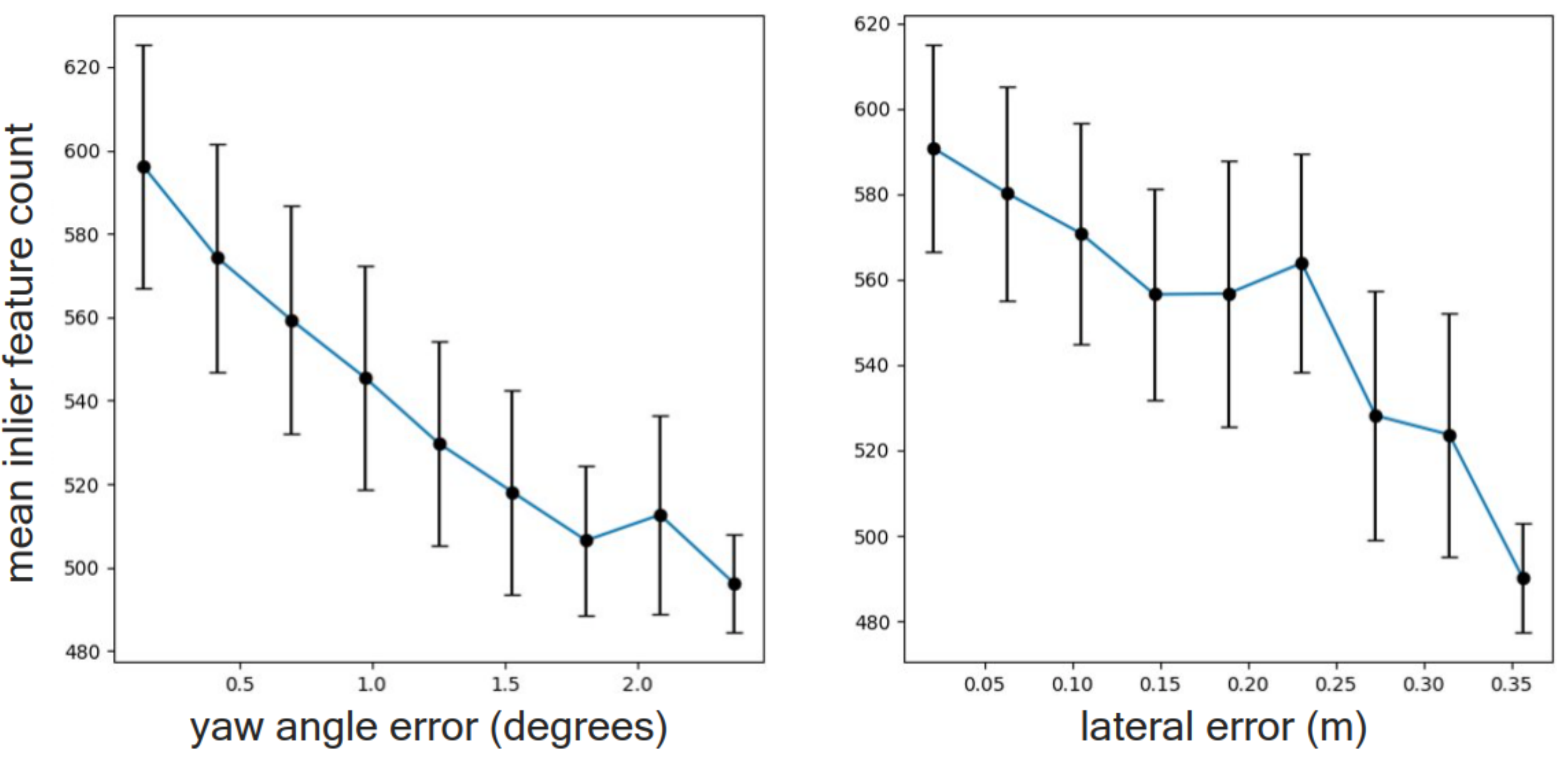}
    \caption{\textbf{Median feature inlier count} (with 1$\sigma$ bound) for different lateral and yaw angle errors of the UTIAS-In-the-Dark test set. The median feature inlier count is highly correlated with the path-following errors, which serves as a reasonable proxy for localization performance in VT\&R.}
    \label{fig:unsup_unet_rmse_features}
    \end{figure}
\vspace{-0.2cm}
\section{EXPERIMENTS}
\vspace{-0.2cm}
    \begin{figure*}[h!]
        \vspace{+0.27cm}
        \centering
        \begin{subfigure}[]{0.3\textwidth}
            \centering
            \includegraphics[width=\textwidth]{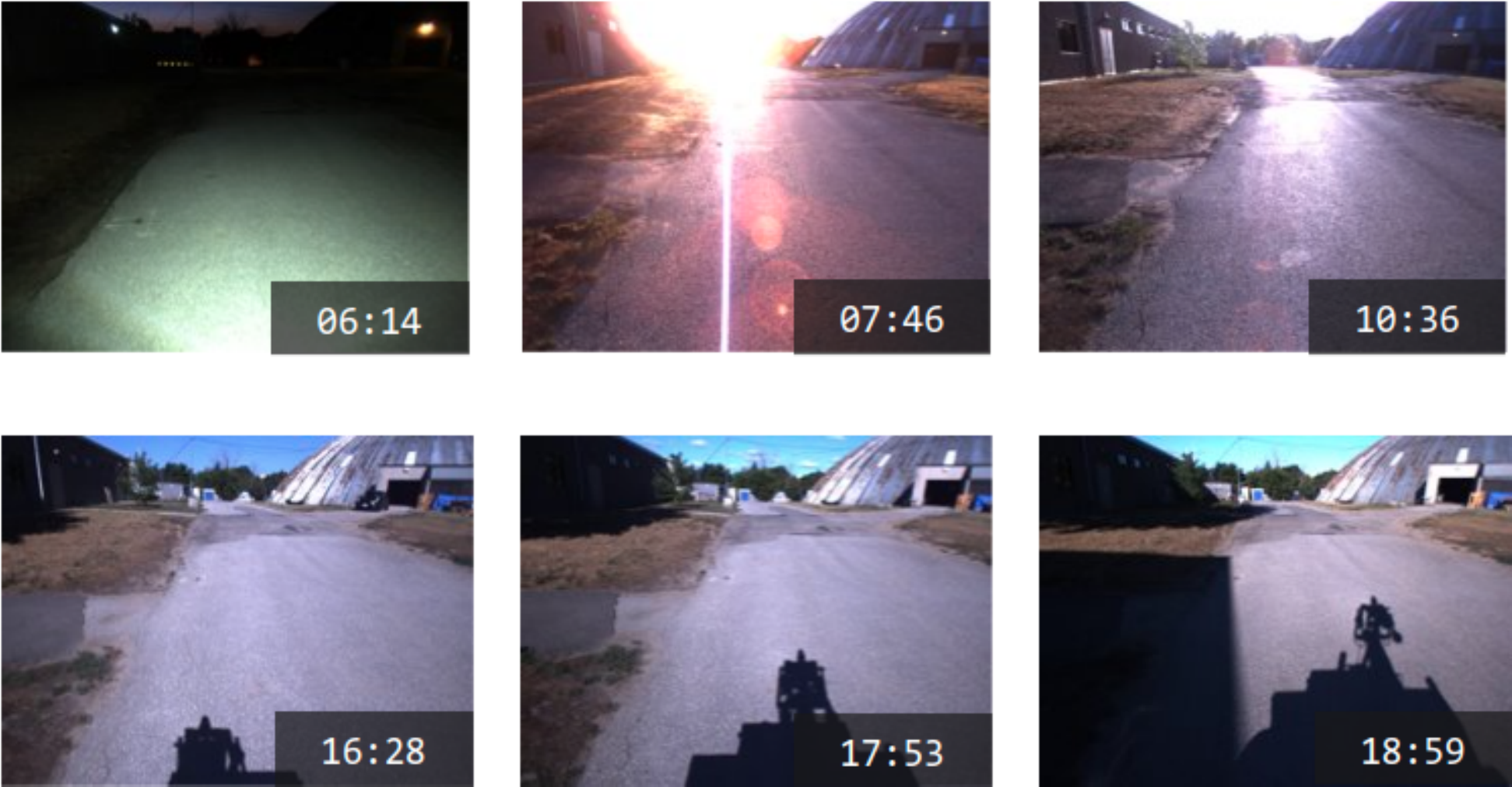}
            \caption{Selected camera views during closed-loop experiment for path 1 (dome).}%
            \label{fig:dome_exp_img}
        \end{subfigure}
        \hfill
        \begin{subfigure}[]{0.3\textwidth}
            \centering 
            \includegraphics[width=\textwidth]{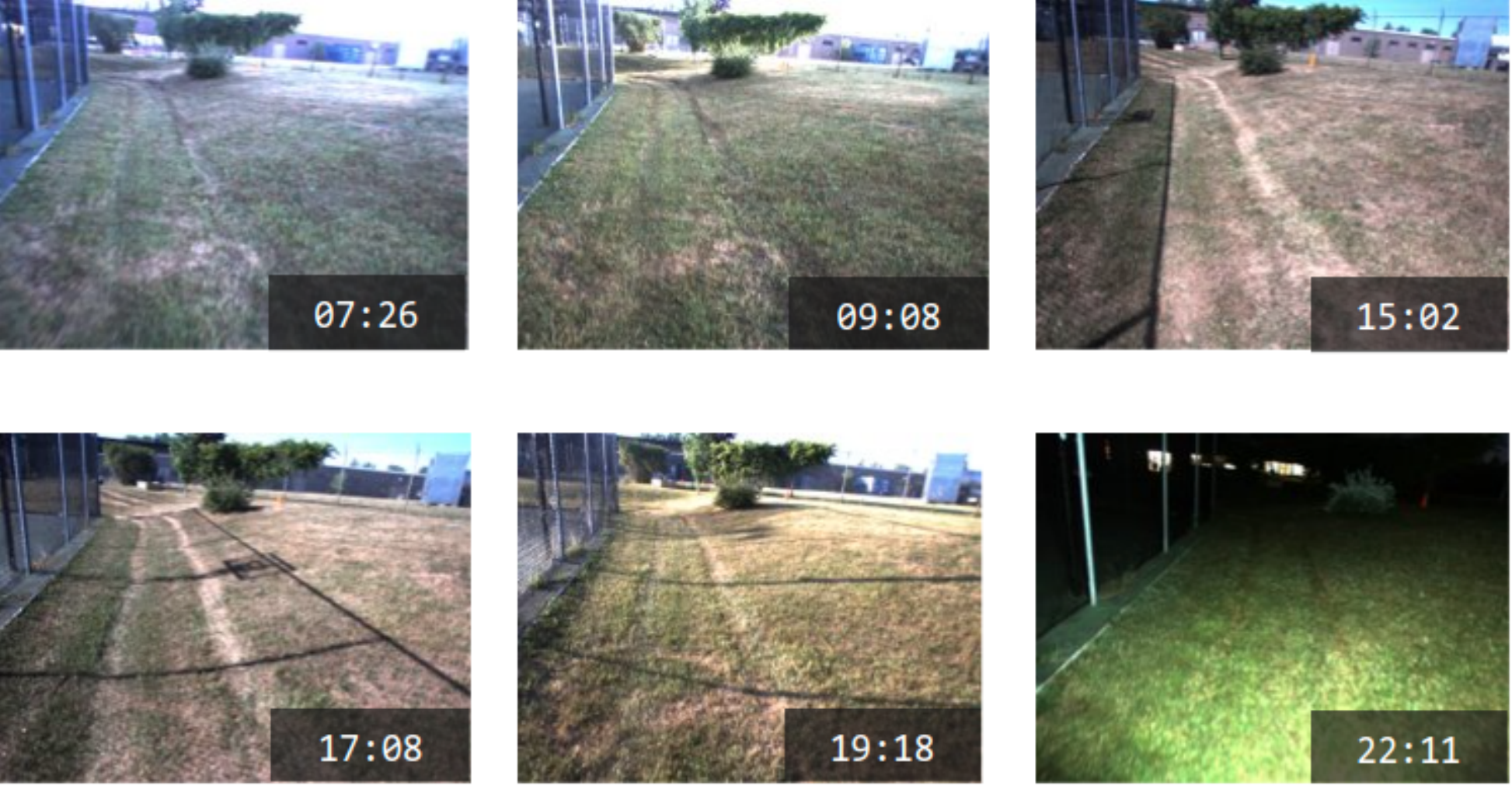}
            \caption{Selected camera views during closed-loop experiment for path 2 (tennis court). }%
            \label{fig:tennis_exp_img}
        \end{subfigure}
        \hfill
        \begin{subfigure}[]{0.3\textwidth}   
            \centering 
            \includegraphics[width=\textwidth]{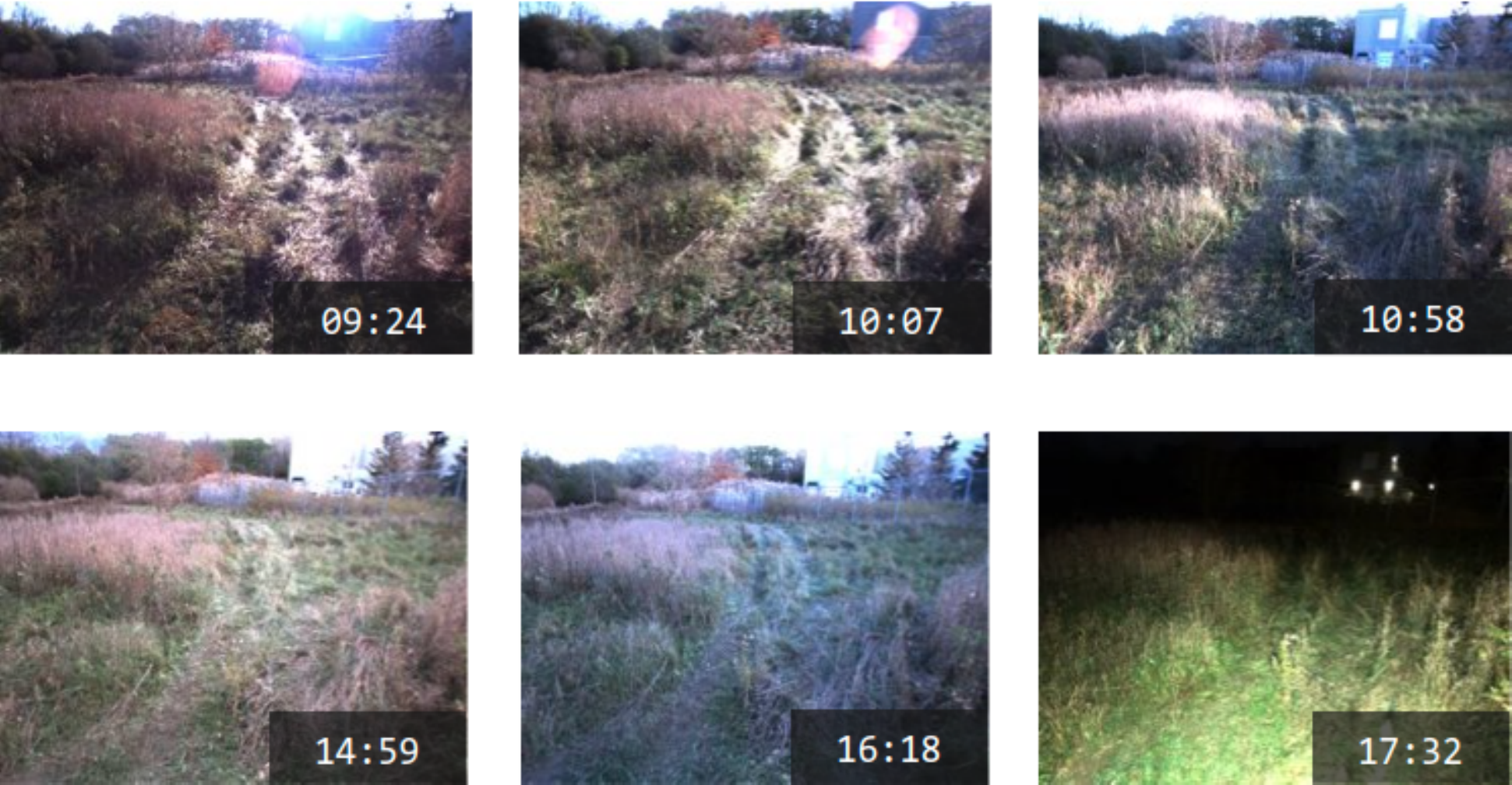}
            \caption{Selected camera views during offline experiment for path 4 (field). }
            \label{fig:field_exp_img}
        \end{subfigure}
        \hspace{0.2cm}
        \caption{\textbf{Selected camera views} from closed-loop and offline experiments labeled with collection time (hh:mm).} 
        \label{fig:selected_image_views}
    \end{figure*}
    
    \begin{table*}[ht]

  \caption{\textbf{Comparison of the path-following errors and the median number of inliers} for SuperPoint\mbox{\cite{superpoint}}, D2-Net\mbox{\cite{D2Net}}, Supervised U-Net\mbox{\cite{mona_vision}}, and our self-supervised method. We construct 6 teach-repeat pairs under different lighting changes for evaluation from the UTIAS-In-the-Dark test set, where we list the experience IDs as well as their collection time for each pair. We report the lateral errors $\Delta y$ in meters, and the yaw angle errors $\Delta \theta$ in degrees.}
  \label{tbl:comparison}
  \centering
\resizebox{\textwidth}{!}{
  \begin{tabu}{lccc|ccc|ccc|ccc|ccc|ccc }
    \cmidrule[\heavyrulewidth]{2-19}

    \multicolumn{1}{c}{}    &
    \multicolumn{3}{c}{\textbf{2-28}} &   
    \multicolumn{3}{c}{\textbf{4-17}} &   
    \multicolumn{3}{c}{\textbf{11-4}} &   
    \multicolumn{3}{c}{\textbf{16-17}} &   
    \multicolumn{3}{c}{\textbf{17-23}} &   
    \multicolumn{3}{c}{\textbf{28-35}}\\
    
    \multicolumn{1}{c}{}    &
    \multicolumn{3}{c}{09:42 - 06:44} &   
    \multicolumn{3}{c}{12:29 - 21:58} &   
    \multicolumn{3}{c}{20:44 - 12:29} &   
    \multicolumn{3}{c}{21:48 - 21:58} &   
    \multicolumn{3}{c}{21:58 - 05:34} &   
    \multicolumn{3}{c}{06:44 - 11:26}\\
    \cmidrule(r){2-19}
    
     &$\Delta y$ & $ \Delta\theta$ & inliers &$\Delta y$ & $\Delta\theta$ & inliers &$\Delta y$ & $\Delta\theta$ & inliers&$\Delta y$ & $\Delta\theta$ & inliers&$\Delta y$ & $\Delta\theta$ & inliers&$\Delta y$ & $\Delta\theta$ & inliers \\
     \midrule
    SuperPoint~\cite{superpoint}&0.18&0.78 &60&0.14&1.2&36&0.22&1.53&72&0.11&0.61&41&0.20&0.89 &51&0.16&1.31&72 \\

    D2-Net~\cite{D2Net} &0.16&1.02&133&0.29&3.68&72&0.05&0.21&278&0.02&0.33&459&0.09&1.29&129&0.18&1.55&122\\
    

    Sup. U-Net~\cite{mona_vision}& \textbf{0.02} &\textbf{0.25} & \textbf{606} &\textbf{0.03} &\textbf{0.47} &\textbf{538} & \textbf{0.03} &0.20 &617 &\textbf{0.02} &\textbf{0.31} & \textbf{614}& \textbf{0.03}& \textbf{0.35}& \textbf{550}& \textbf{0.03}& 0.29 &\textbf{588} \\
     Self Sup. U-Net (Our) & 0.03 & 0.34 &579& 0.06 & 0.50 &536& \textbf{0.03} & \textbf{0.19}&\textbf{619}&\textbf{0.02}&0.32 &611 &0.04 &0.38 &539 &0.04 &\textbf{0.27} &582 \\
    \bottomrule
  \end{tabu}}
  \vspace{-0.3cm}
\end{table*}

\begin{figure}[t]
    \vspace{+0.25cm}
  \centering
  \begin{subfigure}[]{\linewidth}
		\centering        
        \includegraphics[width=\textwidth]{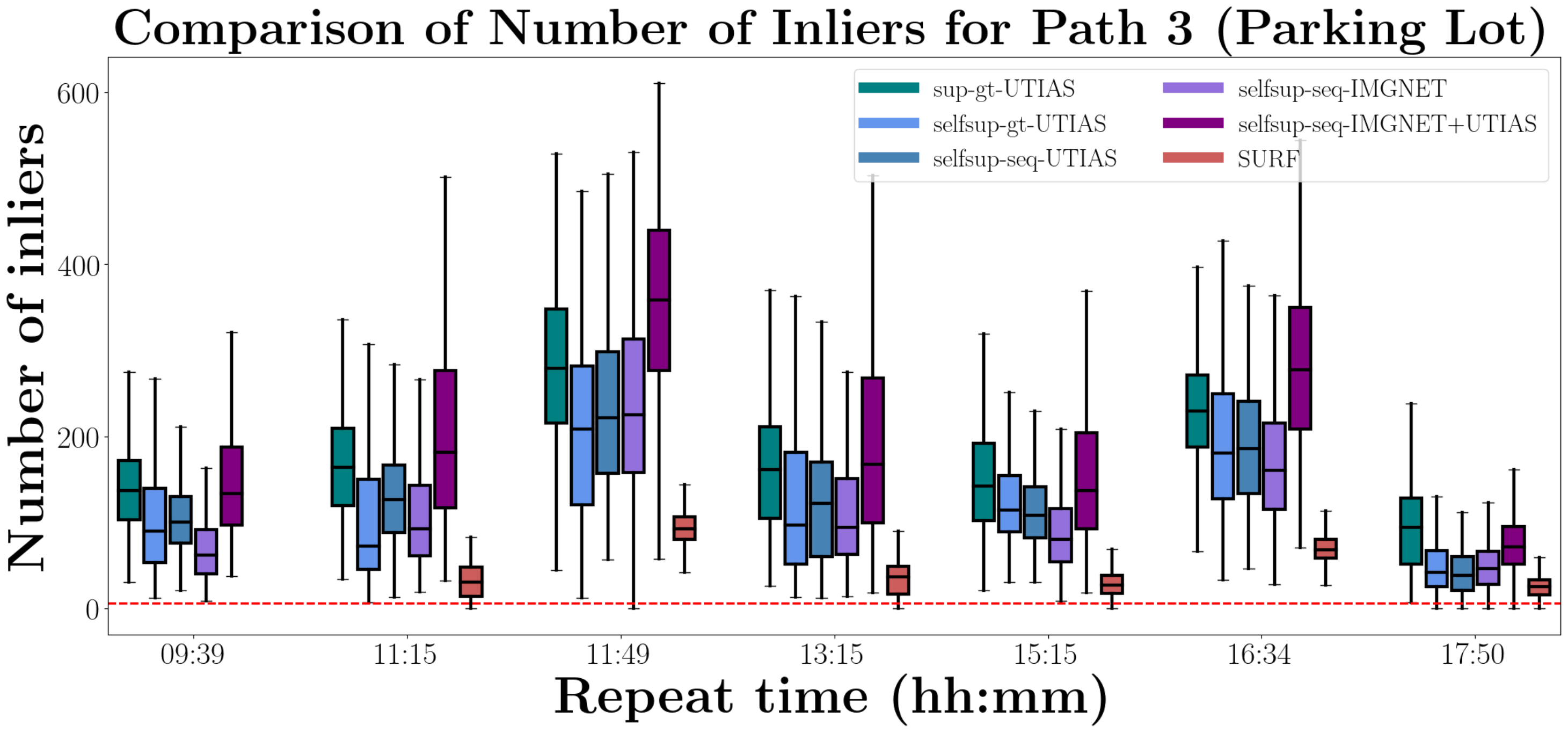}
        \caption{Plot of feature inliers for path 3 (parking lot).}
  \end{subfigure}
  \begin{subfigure}[]{\linewidth}
      \vspace{0.25cm}
  		\centering
        \includegraphics[width=\textwidth]{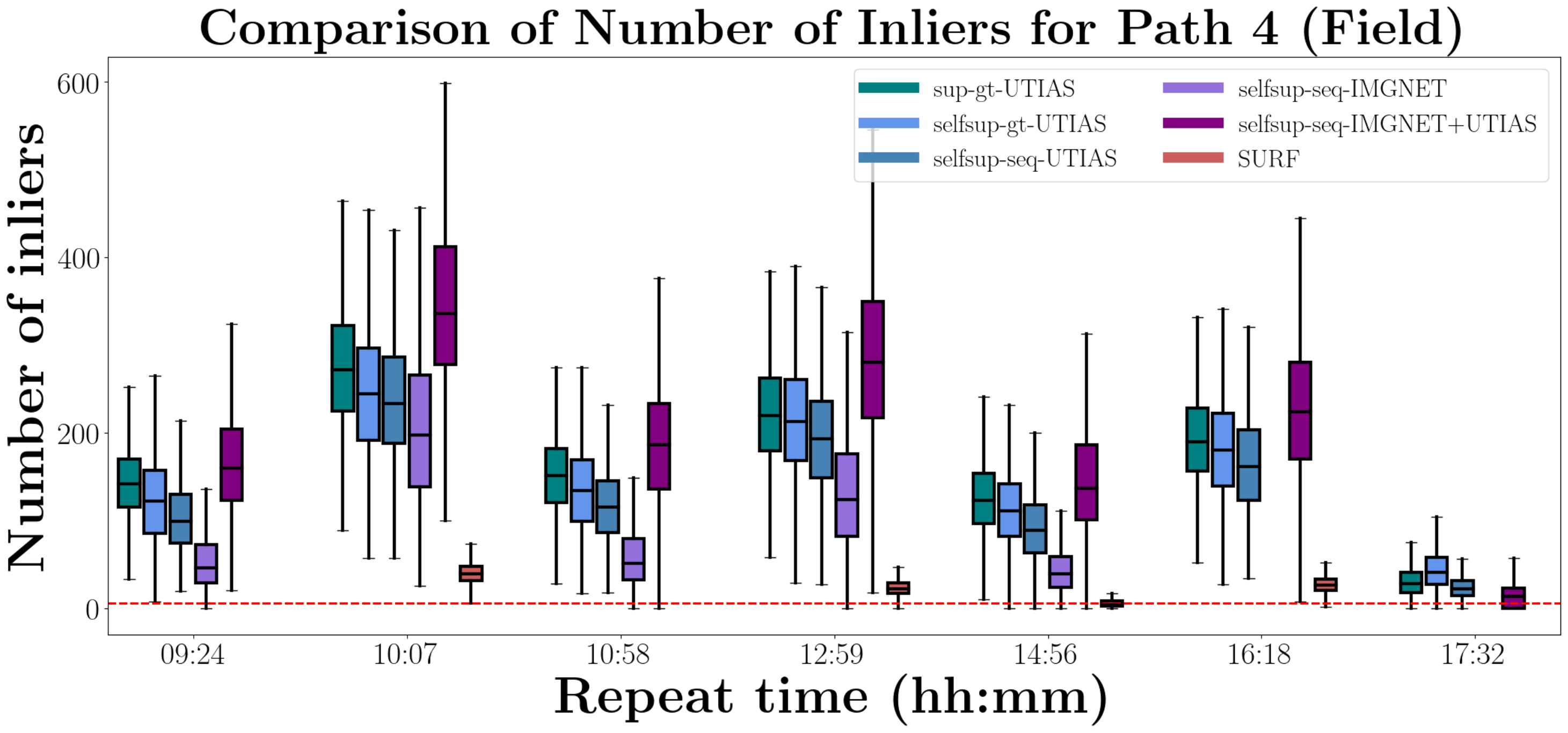}
        \caption{Plot of feature inliers for path 4 (field).}
    \label{fig:offline_feature_inliers}
  \end{subfigure}

  \caption{\textbf{The mean number of inliers for each repeat of the offline experiments} using pre-collected rosbag files for paths 3 (a) and path 4 (b), where `\textit{sup/selfsup}' indicates if the model is trained using supervised or self-supervised losses, `\textit{gt/seq}' indicates if the image pairs are generated using ground-truth poses or SeqSLAM-based algorithm, and `\textit{IMGNET/UTIAS}' represents the datasets that the model is pretrained or finetuned on. A missing entry indicates failure in localization. The learned features outperform SURF in all experiments. Under the exact same settings (i.e., trained on UTIAS datasets only), the self-supervised method achieves competitive results in comparison with the supervised method in all experiments. In addition, the performance increases for the self-supervised method when pretrained on ImageNet and finetuned on UTIAS datasets.}
  \label{fig:ablation_study}
  \vspace{-0.2cm}
\end{figure}

\begin{figure}[h]
    \vspace{+0.25cm}

  \centering
  \begin{subfigure}[]{0.93\linewidth}
  		\centering
        \includegraphics[width=\textwidth]{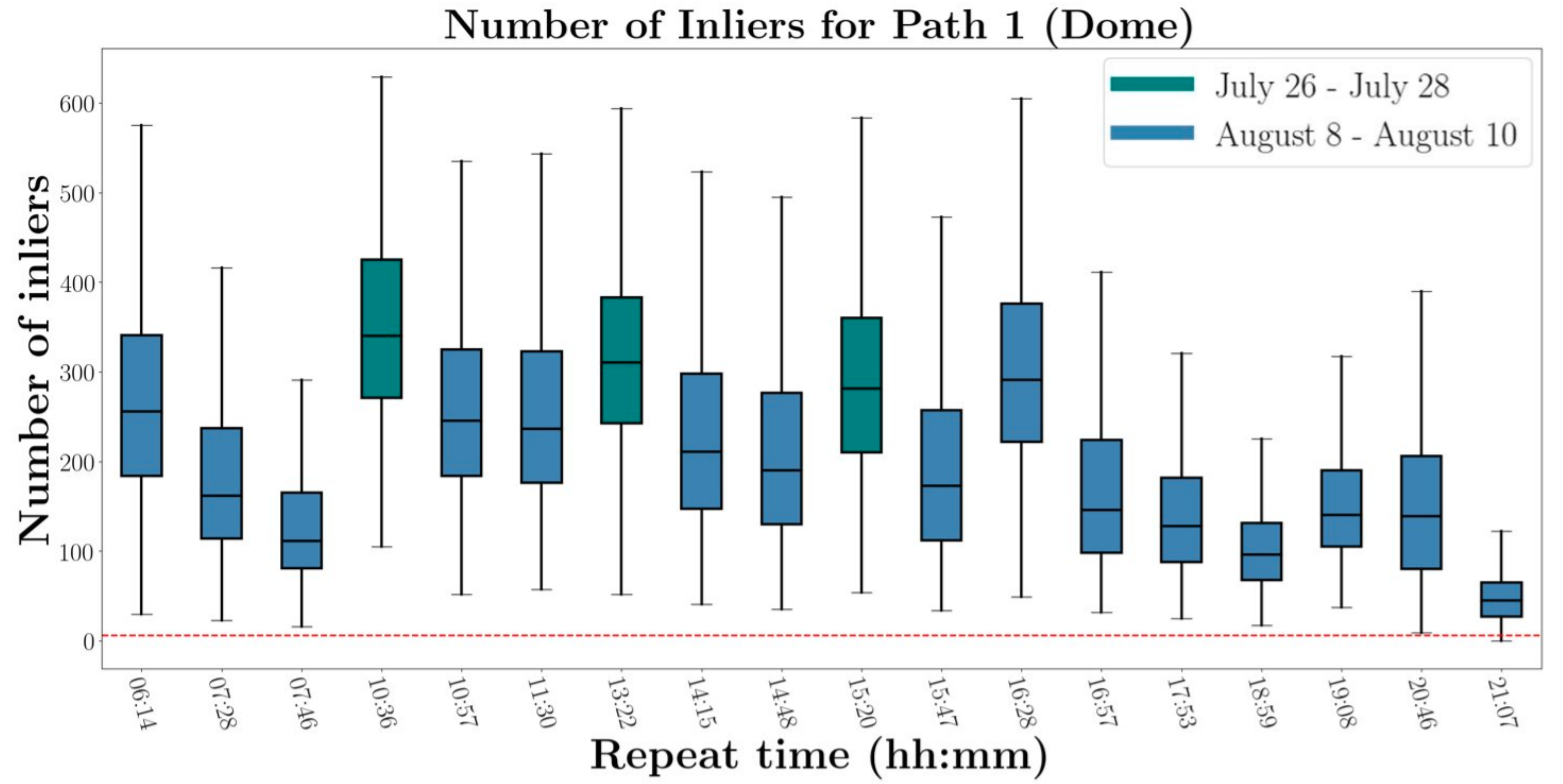}
        \caption{Plot of feature inliers for path 1 (tennis court).}
  \end{subfigure}
  \begin{subfigure}[]{0.92\linewidth}
        \vspace{0.25cm}
		\centering        
        \includegraphics[width=\textwidth]{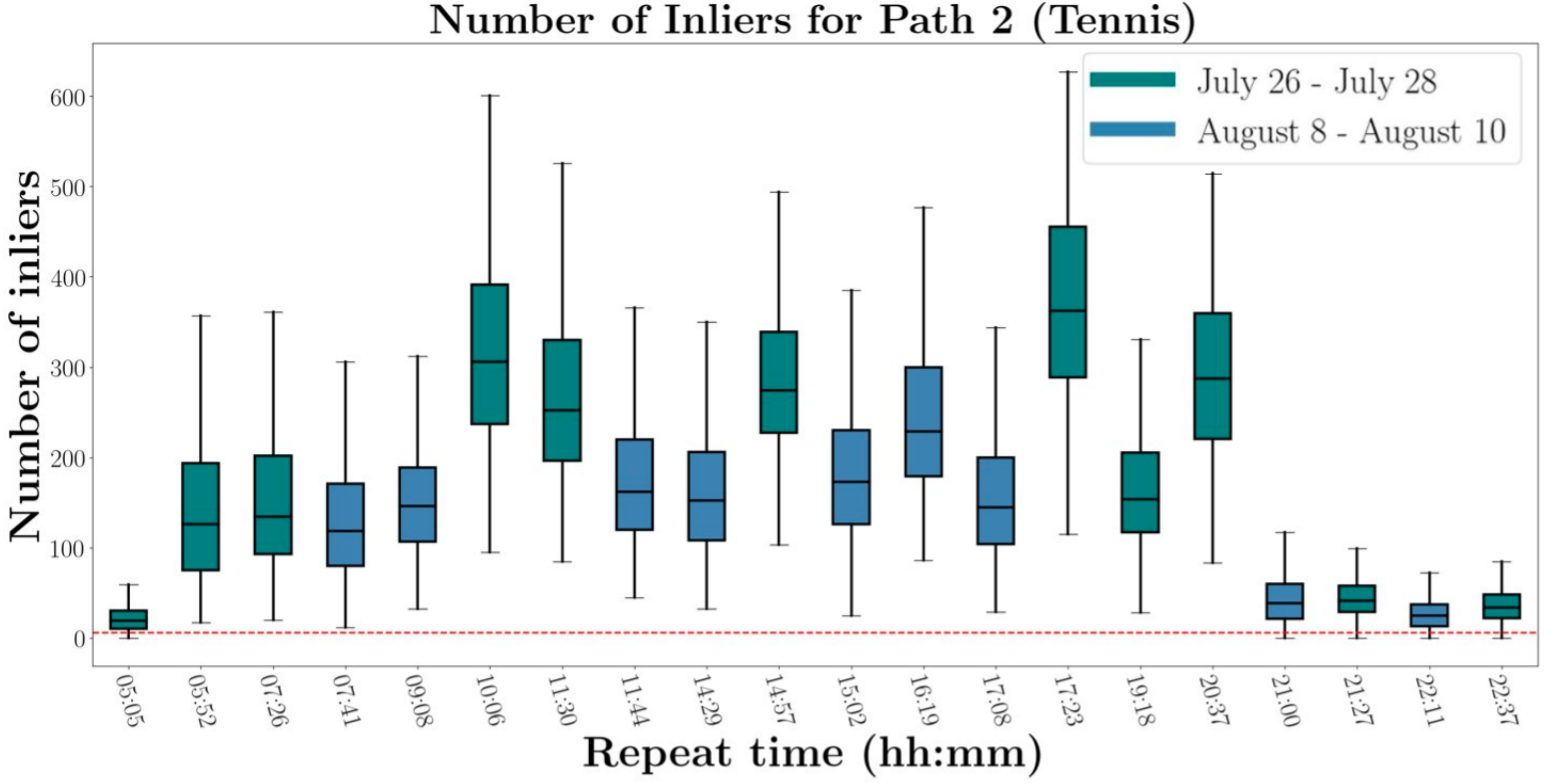}
        \caption{Plot of feature inliers for path 2 (dome).}
  \end{subfigure}

  \begin{subfigure}[]{0.92\linewidth}
    \vspace{0.2cm}
		\centering        
        \includegraphics[width=\textwidth]{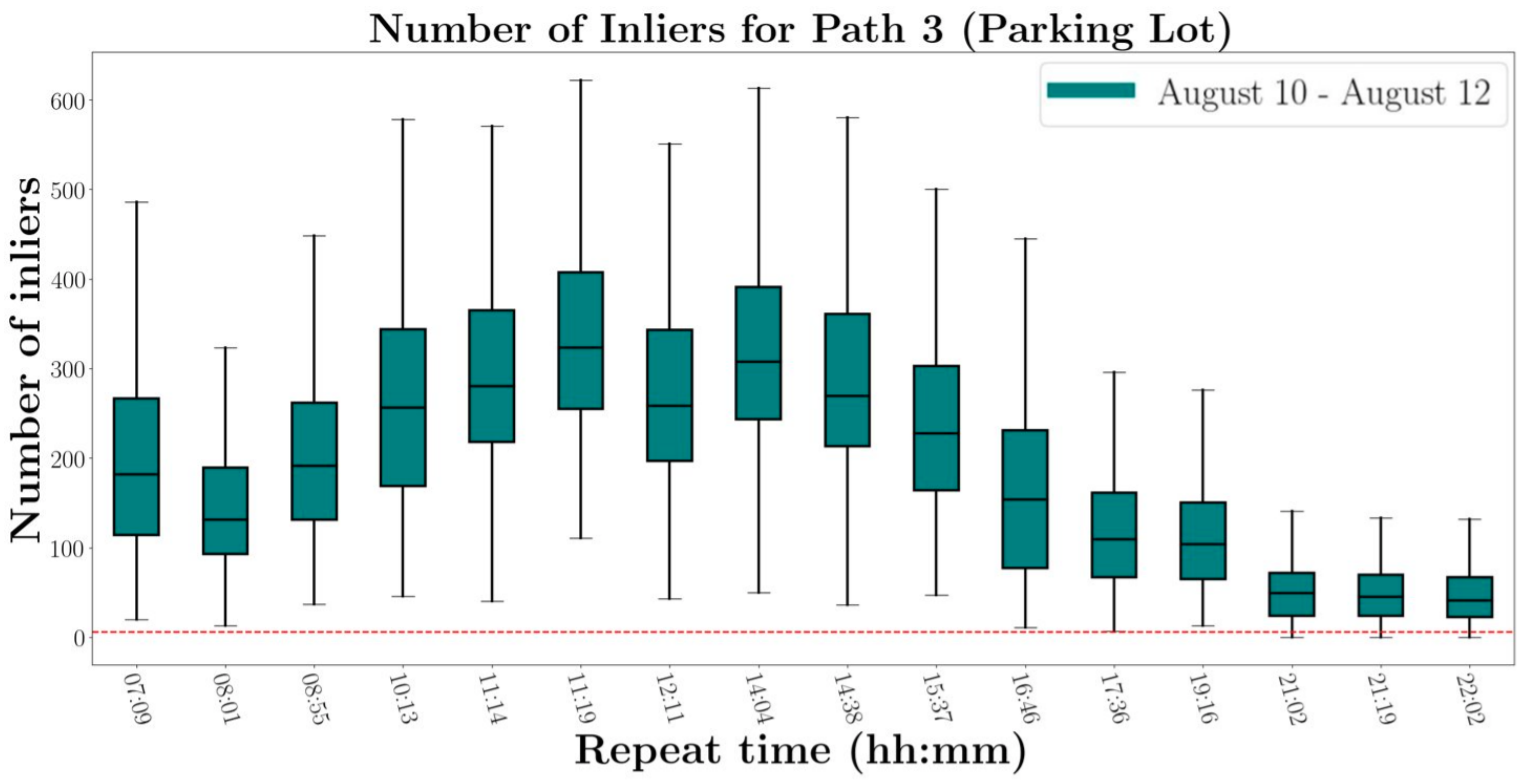}
        \caption{Plot of feature inliers for path 3 (parking lot).}
  \end{subfigure}
  \caption{\textbf{The mean number of inliers for each repeat of the closed loop experiments} for path 1 (a), path 2 (b), and path 3 (c). The red line indicates the minimum number of inliers required for successful localization. The learned features can be integrated with VT\&R to successfully localize under different lighting conditions. In addition, the learned features can be generalized to new regions outside the training data. }
  \label{fig:online_inlier_plots}
  \vspace{-0.2cm}
\end{figure}

\subsection{Datasets}
\vspace{-0.1cm}

For training, we use the same datasets as described in~\cite{mona_vision} to test localization for a route-following problem. The training data were collected using a Clearpath Grizzly robot with a Bumblebee XB3 camera. The robot autonomously repeats two different paths across drastic lighting and seasonal change using Multi-Experience VT\&R~\cite{MEL}. VT\&R stores stereo image keyframes as vertices in a spatio-temporal pose graph, where each edge contains the relative pose between a repeat vertex and the teach vertex. Unlike \cite{mona_vision}, we do not rely on the spatio-temporal pose graph to extract ground-truth relative transforms between vertices for sampling image pairs or for training. 

We use data from two different paths for training, which are included in the In-the-Dark and Multiseason datasets respectively. The In-The-Dark dataset contains 39 runs of a path collected at the campus of UTIAS in summer 2016. The same path was repeated hourly for 30 hours, which captures incremental lighting changes throughout the day. The Multiseason dataset contains 136 runs of a path in an area with more vegetation. The path was consistently repeated from January until May 2017, which captures diverse seasonal and weather conditions throughout the year. Both datasets are publicly available. 

\subsection{Training Data Generation using Topological Localization}
\vspace{-0.1cm}

We generate direct and indirect image correspondences across different sequences in the training datasets as described in Section \ref{methodology}, which are used to sample 100,000 training image pairs and 20,000 validation image pairs. To evaluate the accuracy of the sampled training data, we extract the relative transform between the source and target image from the spatio-temporal graph for all sampled image pairs, then plot the distribution of the absolute distances in Figure \ref{fig:seqslam_training_testing_data}. As can be seen from the histogram, majority of the image pairs are within 3 metres of each other, which generally means sufficient overlapping visual field for training.

\subsection{Training and Inference}
\vspace{-0.1cm}

We use the sampled image pairs to train the network. During training and inference, we discard outlier keypoints using RANSAC. The network is trained using the Adam optimizer with a learning rate of $10^{-5}$ on an NVIDIA Tesla V100 DGXS GPU. The encoder is a VGG16 network \cite{vgg16} pretrained on the ImageNet dataset \cite{IMGNet}, truncated after the $conv\_5\_3$ layer. The network is finetuned end-to-end on the UTIAS Multiseason and In-the-Dark datasets for 50 epochs. 

After training the network, we integrate the learned keypoint detector and descriptor with the VT\&R system to extract features for visual localization. While the user manually drives the robot to teach a path, VT\&R creates a spatio-temporal pose graph that stores relative poses between keyframes. The path is autonomously repeated by computing VO and localizing keyframes. VT\&R relies on SURF for VO, and utilizes the learned features along with a sparse descriptor matcher for localization. 

We conduct three experiments to verify the localization performance of the learned features. We compare our methods to other existing learned features under lighting changes by calculating the path-following errors using the UTIAS-In-the-Dark test set.

In addition, we perform an offline localization ablation study using pre-collected rosbags\mbox{\cite{mona_vision}} for paths 3 and 4. Lastly, we perform closed-loop experiments on three paths (path 1, path 2, and path 3) under different lighting conditions as shown in Figure~{\ref{fig:grizzly}}. We set the maximum speed of the robot to 0.6 m/s. Selected camera views for localization experiments are shown in Figure~{\ref{fig:selected_image_views}}. For these two experiments, it was often not feasible to collect accurate RTK GPS ground truth as we drove in area where tree cover blocks the GPS signals. Hence, we report the median number of inliers in feature matching to reflect the performance of localization. We plot the the median number of inliers against the path-following errors in both lateral and yaw directions using the UTIAS-In-the-Dark test set in Figure~{\ref{fig:unsup_unet_rmse_features}}, which indicates that the median number of inliers is highly correlated with the path-following performance, showing that a higher inlier count is beneficial for accurate and robust localization.


\subsection{Offline Localization: Learned Features Comparison}
\vspace{-0.1cm}

We compared the localization performance of our method to its supervised counterpart (Sup. U-Net)\mbox{\cite{mona_vision}}, SuperPoint\mbox{\cite{superpoint}} and D2-Net\mbox{\cite{D2Net}} in PyTorch. The pretrained networks provided by the authors are used to extract the keypoints, descriptors, and scores. We compute the path-following errors between six teach-repeat pairs sampled from the UTIAS-In-the-Dark test set, then report the lateral errors, yaw angle errors, and the median number of inliers in Table~{\ref{tbl:comparison}}, where the ground truth poses are obtained from Multi-Experience Localization during data collection. Our method is competitive with its supervised counterpart, and outperforms SuperPoint and D2-Net in all metrics for all six localization experiments.

\subsection{Offline Localization: Ablation Study}
\vspace{-0.1cm}

We perform offline experiments on paths 3 and 4 using pre-collected rosbag files, and present the results in Figure~\ref{fig:ablation_study}. Each path is played back 7 times from morning to nighttime with 100\% localization success rate. In the same plot, we compare with different methods. In all experiments, SURF results in a significantly lower number of feature inliers, and fails to localize at the beginning of three different repeats. This shows that the learned features are superior in terms of handling illumination changes. 

When trained on UTIAS datasets only (i.e., not pretrained on ImageNet), we see a slight performance drop in the self-supervised method when compared to the supervised counterpart. Although SeqSLAM matching is not perfect, using SeqSLAM matching to generate training pairs results in a competitive performance compared to using a ground-truth image matches. This suggests the effectiveness of using a place-recognition algorithm to find coarse image correspondences for detailed metric feature learning. 

When trained on ImageNet only, the self-supervised method does not perform well compared to other methods, and failed to localize in two runs for path 4. However, the best performance for the self-supervised method is achieved when pretrained on ImageNet and finetuned on UTIAS datasets, which even outperforms the supervised method (i.e., not pretrained on ImageNet). It shows that by training on UTIAS datasets, it allows the model to generalize better to unseen data. 
    
\subsection{Online VT\&R: Closed-Loop Experiments}
\vspace{-0.1cm}

We perform closed-loop experiments using online VT\&R on path 1, 2, and 3 from Figure \ref{fig:training_testing_path} with 100\% success rate. Path 1 (dome) and path 2 (tennis court) is repeated 18 and 20 times respectively, from 6 am to 9 pm, and from July 26 to August 10 in 2022. Path 3 (parking lot) is repeated 17 times from 7 am to 10 pm in mid-August in 2022. The feature inliers for each repeat is shown in Figure~\ref{fig:online_inlier_plots}. For every repeat, we obtained sufficient feature inliers for localization. In general, we get the highest number of inliers in the middle of the day or when the weather is cloudy, since there are no shadows on the ground. The number of inliers decreases at dusk or dawn, and is at the lowest during the night. In addition, we test on paths that are not within the training data to evaluate the generalizability of the network. Although part of path 1 and path 3 are outside of the training data, we still obtained sufficient number of feature inliers, which shows that the learned features can generalize well to unseen regions.

\section{Conclusion}
\vspace{-0.1cm}
In this paper, we have shown that we can generate image correspondences for image-pair sampling and perform self-supervised feature learning without any ground-truth pose information. We validated the effectiveness of learned features in the VT\&R framework on unseen paths under various lighting conditions. In addition, our pipeline can be readily deployed on self-collected image sequences without additional ground truth data. Limitations of the paper are the assumptions that we made to generate image correspondences. In the future, we intend to further relax these constraints and match sequences of image data that might not start or end at the same place. In addition, we plan to run closed-loop experiments over a longer period of time to test the robustness of the learned features against seasonal change.








{\small
\bibliographystyle{IEEEtran}
\bibliography{egbib}
}


\end{document}